\documentclass[11pt]{article}
\usepackage{acl}
\usepackage{times}
\usepackage{latexsym}
\usepackage[T1]{fontenc}
\usepackage[utf8]{inputenc}
\usepackage{microtype}
\usepackage{inconsolata}
\usepackage{graphicx}
\usepackage{amsmath}
\usepackage{amssymb}
\usepackage{booktabs}
\usepackage{array}
\usepackage{multirow}
\usepackage{enumitem}
\usepackage{tabularx}
\usepackage{adjustbox}
\usepackage{placeins}
\usepackage{url}
\usepackage{xcolor}
\usepackage{hyperref}
\definecolor{ACLLinkBlue}{RGB}{0,0,128}
\hypersetup{colorlinks=true,linkcolor=ACLLinkBlue,citecolor=ACLLinkBlue,urlcolor=ACLLinkBlue}
\providecommand{\tightlist}{\setlength{\itemsep}{0pt}\setlength{\parskip}{0pt}}
\makeatletter
\g@addto@macro\normalsize{%
  \setlength{\abovedisplayskip}{1pt plus 0.5pt minus 0.5pt}%
  \setlength{\belowdisplayskip}{1pt plus 0.5pt minus 0.5pt}%
  \setlength{\abovedisplayshortskip}{0pt plus 0.5pt minus 0pt}%
  \setlength{\belowdisplayshortskip}{1pt plus 0.5pt minus 0.5pt}%
}
\makeatother
\newenvironment{tightdisplay}{%
  \par\nobreak\vspace{4pt}%
  \noindent\hbox to \linewidth\bgroup\hfil\(\displaystyle
}{%
  \)\hfil\egroup\par\nobreak\vspace{4pt}%
}
\newenvironment{tightdisplayline}{%
  \par\nobreak\vspace{4pt}%
  \noindent\hbox to \linewidth\bgroup\hfil\(\displaystyle
}{%
  \)\hfil\egroup\par\nobreak\vspace{4pt}%
}
\newcommand{\racedagger}{\raisebox{0.55ex}{\setlength{\unitlength}{0.65ex}\begin{picture}(0.8,1.15)\linethickness{0.05ex}\put(0.40,0.00){\line(0,1){1.10}}\put(0.12,0.72){\line(1,0){0.56}}\end{picture}}}
\AtBeginDocument{%
  \setlength{\abovedisplayskip}{1pt plus 0.5pt minus 0.5pt}%
  \setlength{\belowdisplayskip}{1pt plus 0.5pt minus 0.5pt}%
  \setlength{\abovedisplayshortskip}{0pt plus 0.5pt minus 0pt}%
  \setlength{\belowdisplayshortskip}{1pt plus 0.5pt minus 0.5pt}%
}

\title{D2C-Routing: Dimension-to-Composition Evidence Routing for Mixed-Origin AI-Generated Text Detection}

\author{
\textbf{Xin Chen} \quad \textbf{Fuwei Zhang} \quad \textbf{Yiqi Tong}\thanks{Corresponding authors.} \\
\textbf{Wei Guo} \quad \textbf{Yutian Xiao} \quad \textbf{Fuzhen Zhuang}\footnotemark[1] \\
School of Artificial Intelligence, Beihang University, Beijing 100191, China \\
\texttt{\{yqtong,zhuangfuzhen\}@buaa.edu.cn}
}

\begin{document}
\maketitle

\begin{abstract}
AI-generated text detection is commonly framed as a binary document-level judgment about whether a text is human-written or machine-generated. This framing breaks down for mixed-origin writing, where content origin and expression origin may differ. We cast mixed-origin detection as dimension-to-composition source attribution, inferring content origin and expression origin before composing them into four collaboration types. We propose \emph{Dimension-to-Composition Routing} (\textbf{D2C-Routing}), which routes content-side and expression-side evidence to supervised dimension heads before a learned gated composition layer predicts the final label. On MixD2C, a reconstructed split derived from the HART mixed-origin benchmark, our disclosed D2C-Routing-based detector system reaches 0.8603 four-way Avg TPR@1\%FPR, 6.5 points above the same-split RACE-local rerun. Core ablations support the routing design, while error analysis shows that distinguishing AI-content/human-expression from fully AI-generated text remains the hardest boundary. Code is available at \href{https://github.com/bystander563/d2c-routing-artifact}{https://github.com/bystander563/d2c-routing-artifact}.
\end{abstract}

\section{Introduction}\label{introduction}

AI-generated text detection commonly asks whether a document is human-written or machine-generated \citep{wang2024m4gtbench,dugan2024raid,tpr-fpr}. This binary framing is increasingly incomplete for collaborative writing workflows, where authorship can be distributed across layers of a document \citep{saha2025almost,zhang-etal-2024-mixset,llmdetect}. For example, one document may preserve human-originated content while using AI-polished wording, whereas another may contain AI-originated content rewritten by a human. A single scalar AI-likeness score cannot distinguish which source dimension changed.

\begin{figure*}[!t]
\centering
\includegraphics[width=0.92\textwidth]{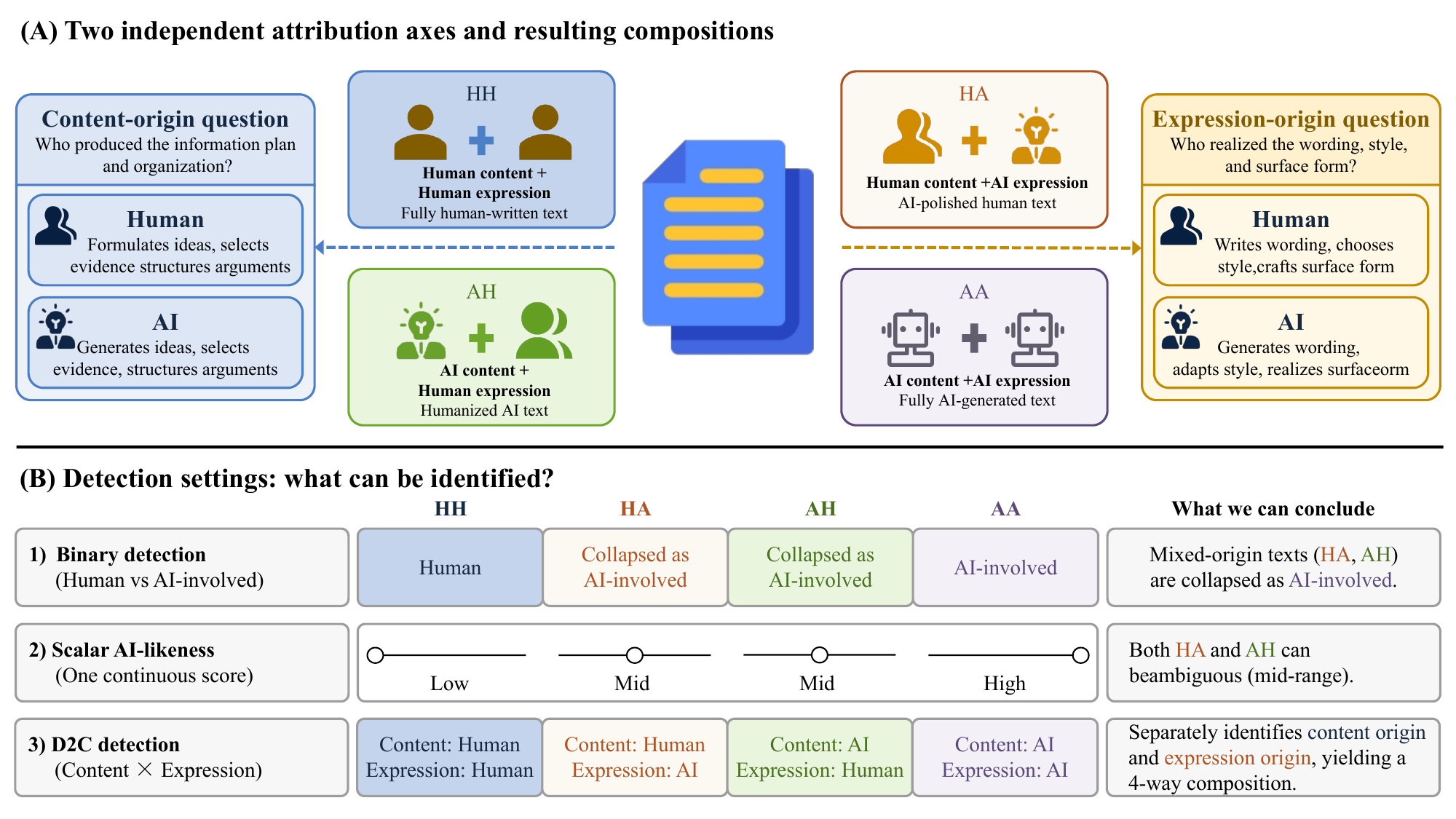}
\caption{Mixed-origin text requires dimension-to-composition detection. We represent attribution through two source dimensions: content origin and expression origin. Their composition yields four labels: HH, HA, AH, and AA. Binary detection collapses mixed-origin cases, while D2C-Routing predicts the two source dimensions before composing the final collaboration type.}
\label{fig:motivation-scope}
\end{figure*}

We adopt the content-expression label space introduced by HART \citep{bao2025-hart}, which defines four mixed-origin collaboration types as the Cartesian product of content origin and expression origin. We use transparent two-letter labels where the first letter denotes content origin and the second denotes expression origin, yielding human-content/human-expression (HH), human-content/AI-expression (HA), AI-content/human-expression (AH), and AI-content/AI-expression (AA). \hyperref[fig:motivation-scope]{Figure~\ref*{fig:motivation-scope}} summarizes the resulting two-axis attribution problem. A binary human-versus-AI detector collapses mixed cases, and a scalar AI-likeness score does not specify which source dimension drives the judgment. We therefore model mixed-origin detection as two source-attribution questions followed by a composed four-way decision. Related representation-learning work has likewise separated semantic and temporal factors before modeling their interaction \citep{zhang2022along}, while multi-aspect alignment provides a way to combine complementary signals before generative prediction \citep{zhang2026multi}.

To address this challenge, we introduce \emph{Dimension-to-Composition Routing} (\textbf{D2C-Routing}), where D2C denotes the path from supervised source dimensions to their composed collaboration label. D2C-Routing organizes document-internal evidence as a dimension-aligned tree. Hierarchical knowledge fusion can preserve evidence at different semantic levels \citep{zhang2022mind}, while multi-level relevance learning can distinguish coarse from fine relationships \citep{zhang2025multi}. Content-side cues such as entity continuity and discourse motifs support the content side, while expression-side cues such as lexical choice, rhythm/POS patterns, and surface regularity support the expression side. The routed evidence then feeds supervised content-origin and expression-origin heads before a learned gate composes HH/HA/AH/AA.

We evaluate on MixD2C, a four-way split reconstructed from the released HART benchmark files. The experiments separate single-model method evidence from detector-system performance and test dimension supervision and learned composition against feature, capacity, ranking, and fusion controls.

This design yields a direct empirical test: the intermediate source dimensions should be learnable, and their composition should improve low-FPR four-way detection.

\noindent\begin{minipage}{\columnwidth}
\hspace*{0.4cm}Our contributions are:

\begin{enumerate}[topsep=2pt,partopsep=0pt,itemsep=0pt,parsep=0pt]
\def\labelenumi{\arabic{enumi}.}
\tightlist
\item
  We organize document-internal evidence into linguistically motivated content and expression branches for mixed-origin AI-text detection.
\item
  We introduce D2C-Routing, which combines dimension-specific pathways, supervised content-origin and expression-origin heads, and learned gated composition into HH/HA/AH/AA collaboration labels.
\item
  Under the MixD2C split, D2C-Routing improves average and AA low-FPR ranking over same-split single-model controls. A D2C-Routing-based detector system reaches 0.8603 four-way Avg TPR@1\%FPR, 6.5 points above the same-split RACE-local rerun.
\end{enumerate}
\end{minipage}
\vfill
\newpage

\section{Related Work}\label{related-work}

AI-generated text detection usually evaluates documents with binary human/AI labels or scalar AI-likeness scores. Recent work broadens this setting to AI-polished, collaborative, or fine-grained writing \citep{saha2025almost,zhang-etal-2024-mixset,ta-etal-2026-faid}, multi-generator and manipulated-text benchmarks \citep{guo2023hc3,wang-etal-2024-m4,wang2024m4gtbench,dugan2024raid,wang-etal-2025-real,pan2025generated}, and finer-grained role, involvement, boundary, or process predictions \citep{llmdetect,roft-boundary,detree}. These studies motivate the same broad concern that a document may not have a single source. They provide important context, but they are not all direct numerical baselines for our main four-way MixD2C table because they differ in label space, prediction granularity, reconstruction protocol, or primary metric.

For the content-expression taxonomy, we use HART's HH/HA/AH/AA label space and official Level-1/2/3 risk tasks \citep{bao2025-hart}. We do not claim to introduce this taxonomy. A closely related HART-family study, RACE, reports a broad baseline set under a 70/10/20 reconstruction \citep{li2026race}. We keep its published table as external context and rerun the released RACE implementation on the identical MixD2C files used by our models. Because the release does not package frozen published sample IDs or checkpoints, the local run is a same-split RACE rerun, not a reproduction of the published RACE numbers.

Training-free scalar detectors include Fast-DetectGPT, which uses conditional probability curvature for efficient zero-shot detection; Binoculars, which contrasts two language models; and SpecDetect, which analyzes token log-likelihood sequences in the frequency domain \citep{bao2024fastdetectgpt,hans2024binoculars,luo2026specdetect}. Recent analysis connects spectral energy to variance in proxy log-probability trajectories and finds that spectral cues weaken for short, fragmented, mixed, and edited text \citep{luo2026spectralmechanisms}. Related scalar-detection work includes GLTR's token-rank visualization, DetectGPT's perturbation-based curvature test, and Ghostbuster's trained feature classifier \citep{gehrmann-etal-2019-gltr,mitchell2023detectgpt,verma-etal-2024-ghostbuster}. Because the three evaluated training-free methods return scalar scores rather than native HH/HA/AH/AA predictions, we evaluate them directly on official binary collapses and report four-way results only as calibrated diagnostics.

Beyond AI-text detection, multi-view and auxiliary-supervision methods provide useful precedents for preserving heterogeneous evidence before a final decision. MindScore decomposes text-to-image preference into matching, faithfulness, quality, and realness \citep{tong2025mindscore}. In NLP, multi-task MRC supervision has been used to integrate structured prior knowledge \citep{tong2022biomrc}, while dynamic caching of inter-sentence representations has improved consistency across local predictions \citep{tong2022biocache}.

Our method differs in its inductive bias. RACE emphasizes creator/editor traces through rhetoric-guided graph learning, while D2C-Routing asks how document-internal evidence should be assigned to supervised source dimensions before the final four-way decision. Entity coherence and discourse organization are motivated by coherence modeling and document-level discourse analysis \citep{barzilay-lapata-2008-local,li-hovy-2014-model,liu-etal-2023-coco,mann1988rhetorical,prasad-etal-2008-penn,liu-etal-2021-dmrst,kim-etal-2024-mfidf}; lexical choice, function-word patterns, rhythm, and surface regularity are motivated by stylometry and AI-generated text style analysis \citep{stamatatos2009survey,uchendu-etal-2020-authorship,soto2024fewshot,reinhart2025llms}. D2C-Routing assigns these evidence families dimension-specific roles and tests them against flat-concatenation and feature-removal controls.

The next section makes this source-dimension view explicit before describing the detector architecture.

\section{Problem Setup}\label{problem-setup}

\subsection{Task Definition}\label{task-definition}

We formalize mixed-origin detection as source-dimension prediction followed by composition. Let 
\begin{tightdisplay}
\begin{aligned}
\mathcal{Y} &= \{HH,HA,AH,AA\}, \\
y &= (y_c,y_e),\quad y_c,y_e \in \{0,1\},
\end{aligned}
\end{tightdisplay}
 where \(y_c\) denotes content origin, \(y_e\) denotes expression origin, 0 denotes human-originated, and 1 denotes AI-originated. A detector maps an observed document \(x\) to \((\hat y_c,\hat y_e,\hat y)\), separating the two source questions from the final collaboration label.

\subsection{Label Space}\label{label-space}

The HART taxonomy assigns each document a four-way type based on content origin and expression origin \citep{bao2025-hart}. We use human and AI as the two source values for each dimension, with two-letter labels defined by content origin followed by expression origin.

\begin{table}[!t]
\centering
\scriptsize
\renewcommand{\arraystretch}{1.04}
\caption{Label space. HH/HA/AH/AA compose content origin and expression origin. The two derived binary targets, content origin and expression origin, supervise D2C-Routing; Level-1/2/3 are official HART evaluation collapses, not additional training labels.}
\label{tab:hart-label-space}
\resizebox{\columnwidth}{!}{%
\begin{tabular}{@{}lllcc@{}}
\toprule
\multicolumn{5}{@{}l}{\textit{Panel A: Labels and targets}} \\
\midrule
Label & Content & Expression & C tgt. & E tgt. \\
\midrule
\textbf{HH} & human & human & 0 & 0 \\
\textbf{HA} & human & AI & 0 & 1 \\
\textbf{AH} & AI & human & 1 & 0 \\
\textbf{AA} & AI & AI & 1 & 1 \\
\addlinespace[0.25em]
\midrule
\multicolumn{5}{@{}l}{\textit{Panel B: Official task collapses}} \\
\midrule
Task & Positive & Negative & \multicolumn{2}{l}{Question} \\
\midrule
\textbf{Level-1} & HA + AH + AA & HH & \multicolumn{2}{l}{Any AI?} \\
\textbf{Level-2} & AH + AA & HH + HA & \multicolumn{2}{l}{AI content?} \\
\textbf{Level-3} & AA & HH + HA + AH & \multicolumn{2}{l}{Fully AI?} \\
\bottomrule
\end{tabular}%
}
\end{table}

This factorization provides explicit supervision for the two source decisions underlying the four-way label.

The two source targets are derived as 
\begin{tightdisplay}
\begin{aligned}
y_c &= \mathbb{I}\{y \in \{AH,AA\}\}, \\
y_e &= \mathbb{I}\{y \in \{HA,AA\}\}.
\end{aligned}
\end{tightdisplay}
 A flat classifier can learn four-way label correlations without identifying which source dimension supports a decision. Dimension supervision instead makes content and expression attribution explicit training targets.

\begin{figure*}[!t]
\centering
\includegraphics[width=0.95\textwidth]{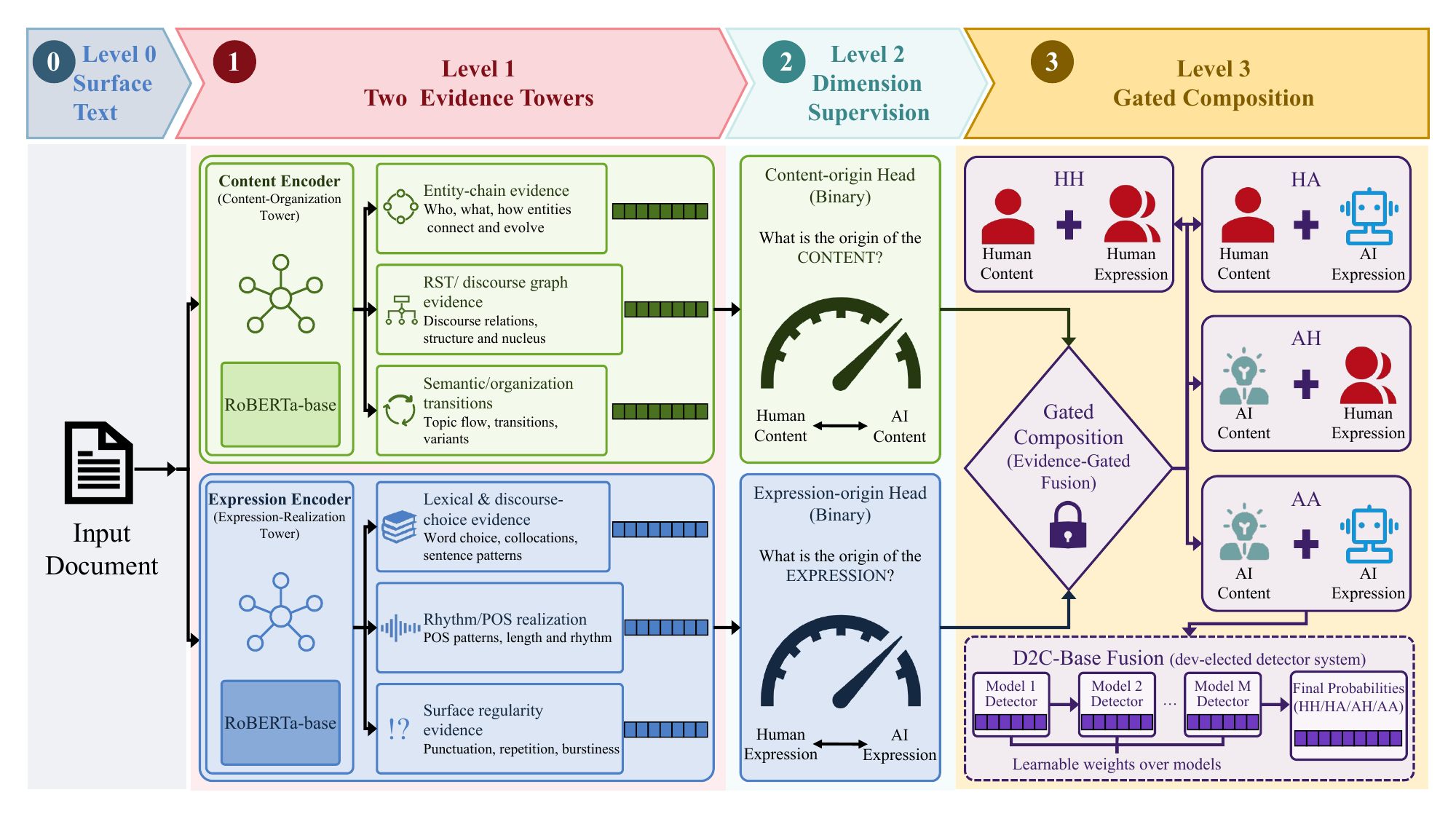}
\caption{D2C-Routing architecture. The figure shows the single-model architecture: the observed document feeds content and expression evidence pathways, supervised source heads estimate the two dimensions, and a learned gate composes them into HH/HA/AH/AA predictions. The separate D2C-base Fusion detector system combines multiple D2C-Routing outputs and is reported only in result tables.}
\label{fig:d2c-architecture}
\end{figure*}

\subsection{Evidence Organization}\label{evidence-organization}

The content-expression decomposition organizes evidence into two linguistically motivated pathways. The content pathway uses entity-chain coherence and RST discourse motifs, while the expression pathway uses lexical-connective choices, rhythm/POS patterns, and surface regularity. The final layer composes the two source dimensions into HH/HA/AH/AA. Syntax-specific routing remains exploratory.

The content and expression predictions are supervised directly, and the four-way prediction is also supervised directly. This dimension-to-composition structure is different from the official Level-1/2/3 tasks, which are evaluation collapses of the same four labels.

The distinction is most important for HA and AH. Under a scalar AI-likeness view, both can appear intermediate between HH and AA. Under a dimension-to-composition view, they are structurally different because HA has human content but AI expression, whereas AH has AI content but human expression.

\section{Method}\label{method}

\subsection{Architecture Overview}\label{architecture-overview}

D2C-Routing instantiates the dimension-aligned evidence tree in \hyperref[fig:d2c-architecture]{Figure~\ref*{fig:d2c-architecture}}. The key design choice is not simply to add more features to a text encoder, but to route document-internal evidence through supervised source-dimension pathways before the final four-way composition. A RoBERTa encoder first produces the observed-document representation \citep{roberta}; content components extract entity-chain and discourse-motif evidence; expression components extract lexical-connective realization, rhythm/POS, and surface-regularity evidence; and side-specific fusion modules form content-origin and expression-origin pathway states before gated composition.

This routing is an inductive bias over document-internal evidence, not a claim that evidence families are causally isolated. Each pathway therefore keeps a linguistic rationale for its inputs while allowing the learned composition module to use entangled evidence when it helps prediction.

The same design can be instantiated with a shared encoder or with separate encoders for the content and expression pathways. Dual-encoder variants use the same encoder family but a larger parameter budget, so they are treated as supportive system variants rather than capacity-matched proof of architecture superiority.

\subsection{Document Encoder}\label{document-encoder}

The raw document is the only directly observed object, so both source dimensions must remain grounded in the same text. Given an input document \(x\), the observed-document layer encodes the original text and produces a contextual representation. Shared-encoder models use the same representation for both sides,

\begin{tightdisplayline}
h_c = h_e = h.
\end{tightdisplayline}

Dual-encoder models instead use separate encoders to produce \(h_c\) and \(h_e\); their larger parameter budget is evaluated separately in the matched system controls.

\subsection{Content Pathway}\label{content-pathway}

The content pathway models document-internal evidence about how information is maintained and organized. Rather than verifying factual correctness, it captures proxies for information planning. We use entity-chain evidence for referent recurrence, sentence overlap, chain length, and local entity transitions, following coherence modeling and coherence-based machine-text detection \citep{barzilay-lapata-2008-local,liu-etal-2023-coco}. We also use RST discourse evidence for relation and motif features, motivated by Rhetorical Structure Theory and discourse-motif detection \citep{mann1988rhetorical,liu-etal-2021-dmrst,kim-etal-2024-mfidf}. Learned projectors map the entity vector \(a_{\mathrm{ent}}(x)\) and discourse vector \(a_{\mathrm{rst}}(x)\) into the content pathway; the entity-as-text variant adds an encoded entity-text vector as a third content component. The final content-origin state concatenates the observed-document anchor \(h_c\) with routed content evidence and predicts whether the document's content is AI-originated.

\subsection{Expression Pathway}\label{expression-pathway}

The expression pathway models how information is linguistically realized. This branch is needed because polishing or humanization can preserve organization while changing vocabulary, connectives, rhythm, grammatical realization, punctuation, repetition, and burstiness. We group these signals into lexical-connective realization \(a_{\mathrm{lex}}(x)\), rhythm/POS realization \(a_{\mathrm{rhy}}(x)\), and surface regularity \(a_{\mathrm{reg}}(x)\), drawing on stylometry and AI-generated text style analysis \citep{stamatatos2009survey,soto2024fewshot,reinhart2025llms}. Learned projectors map each component into the expression pathway, and the final expression-origin state concatenates \(h_e\) with routed expression evidence. This pathway is supervised to predict whether wording and surface realization are AI-originated.

\subsection{Dimension Heads}\label{dimension-heads}

Because the four-way label factorizes into content-origin and expression-origin dimensions, we derive two supervision targets from the same label space and implement binary heads over the routed pathway states. The content head \(s_c = W_c z_c + b_c\) predicts AI-originated content, with AH/AA as positive labels and HH/HA as negative labels. The expression head \(s_e = W_e z_e + b_e\) predicts AI-originated expression, with HA/AA as positive labels and HH/AH as negative labels. Both heads receive direct cross-entropy supervision.

\subsection{Gated Composition}\label{gated-composition}

The final collaboration type is the composition of the two source dimensions, but the learned dimension representations are uncertain and partly entangled. A hard mapping from two binary decisions to four labels would discard this uncertainty, while a flat classifier need not expose or preserve these source dimensions. D2C-Routing therefore projects the pathway representations into a shared composition space, 
\begin{tightdisplayline}
u_c = f_c(z_c), \qquad u_e = f_e(z_e).
\end{tightdisplayline}
 Let \(h_g\) denote the text anchor used by the composition gate: in shared-encoder models, \(h_g=h\); in dual-encoder models, \(h_g\) is a projected combination of \(h_c\) and \(h_e\). The learned vector gate, fused representation, and four-way classifier are 
\begin{tightdisplay}
\begin{aligned}
g &= \sigma(W_g [h_g; s_c; s_e] + b_g),\\
u &= g \odot u_c + (1-g) \odot u_e,\\
p(y \mid x) &= \mathrm{softmax}(W_y [h_g; u; s_c; s_e] + b_y).
\end{aligned}
\end{tightdisplay}

This design differs from a hard factorized classifier that maps two binary predictions directly to four labels. The gate allows the model to learn how content-origin and expression-origin evidence interact across HH, HA, AH, and AA.

\subsection{Training Objective}\label{training-objective}

The training objective mirrors the hierarchy: dimension losses teach the two source questions, the composition loss teaches HH/HA/AH/AA, and low-FPR ranking terms adapt the detector to strict false-positive operating points,

\begin{tightdisplay}
\setlength{\jot}{2pt}
\begin{aligned}
\mathcal{L} ={}& \lambda_c \mathrm{CE}(s_c, y_c) + \lambda_e \mathrm{CE}(s_e, y_e) + \lambda_m \mathcal{L}_{AH/AA} \\
&+ \lambda_y \mathrm{CE}(p(y \mid x), y) + \lambda_r \mathcal{L}_{AA}.
\end{aligned}
\end{tightdisplay}

The first two terms supervise the content and expression dimensions. The third separates AI-content/human-expression from fully AI-generated text, the fourth supervises the composed HH/HA/AH/AA label, and the final AA one-vs-rest ranking term supports low-FPR AA retrieval. The ranking terms are evaluation-aligned optimization choices motivated by low-FPR detector evaluation \citep{tpr-fpr}.

\subsection{Detector-System Fusion}\label{detector-system-fusion}

D2C-base Fusion is the final detector-system variant for low-FPR evaluation. It combines member models that each produce a four-way probability vector \(p_m(y \mid x)\). The candidate pool and development metric are fixed before test evaluation; interpolation weights are selected on development and evaluated once on test, with no test-set interpolation tuning.

\begin{tightdisplayline}
p_{\mathrm{fusion}}(y \mid x) = \sum_m \alpha_m p_m(y \mid x),\; \sum_m \alpha_m = 1.
\end{tightdisplayline}

Official Level-1/2/3 scores are computed by collapsing the fused four-way probabilities into the corresponding positive sets. Single-model architecture evidence is evaluated separately through dimension-head, feature, and composition controls.

\section{Experimental Setup}\label{experimental-setup}

\subsection{Dataset Split}\label{dataset-split}

Because the released benchmark files do not provide frozen RACE sample identifiers, we construct MixD2C as a transparent four-way evaluation split, not a new dataset or taxonomy. MixD2C merges the released development and test JSON files, then stratifies by domain and class into a 70/10/20 train/dev/test split. The split follows the counts reported in the RACE appendix table rather than the 70/20/10 ratio stated in the RACE prose. MixD2C contains 11,200/1,600/3,200 train/dev/test examples, with AH as the minority class; full class counts are reported in \hyperref[mixd2c-split]{Appendix~\ref*{mixd2c-split}}.

The four-way comparison has two RACE-related sources. Published RACE Table 2 values are external references kept for audit context in \hyperref[mixd2c-split]{Appendix~\ref*{mixd2c-split}}. The main same-split comparison uses our run of the released RACE implementation on the identical MixD2C train/dev/test files, with the strict rstdt RST parser setting. This local rerun is same-split with our models, but it is not a reproduction of the published RACE numbers because the release does not include frozen published sample IDs or checkpoints.

\subsection{Metrics}\label{metrics}

For official Level-1/2/3 task collapses, we report AUROC, F1-score, and TPR@5\%FPR. For fine-grained HH/HA/AH/AA diagnostics, AUROC and F1-score are macro-averaged unless stated otherwise; we also report class-wise TPR@1\%FPR and their average. These low-FPR metrics reflect detector evaluation under strict false-positive constraints \citep{tpr-fpr}.

\subsection{Baselines}\label{baselines}

We compare against four protocol-separated baseline groups. First, RACE-local is our rerun of the released RACE implementation on the identical MixD2C split. Second, the main four-way table restores representative published RACE Table 2 references---RoBERTa, RoBERTa-DANN, CoCo, LF-Motifs, and RACE---as related-reconstruction context rather than local reruns \citep{li2026race}; the complete 13-row reference set remains in \hyperref[mixd2c-split]{Appendix~\ref*{mixd2c-split}}. Third, same-split internal controls compare text-only RoBERTa, flat feature concatenation, D2C-Routing, and closely cost-matched ensembles. Fourth, training-free scalar detectors, including SpecDetect, Fast-DetectGPT, and Binoculars, are evaluated on official Level-1/2/3 collapses because their scalar outputs are not native HH/HA/AH/AA predictions. All same-split rows use the same MixD2C files and evaluator; original HART-protocol references remain separate in \hyperref[original-hart-protocol-references]{Appendix~\ref*{original-hart-protocol-references}}.

Table names follow a fixed convention. RACE-local denotes the same-split RACE rerun. D2C-Routing denotes independently evaluated single models; parenthetical tags such as shared or dual identify the encoder instantiation. D2C-base Fusion is the development-selected primary detector system, a probability fusion over RoBERTa-base-family member models. Supplementary fusion variants appear only in robustness and appendix analyses. The strongest single-model row is a dual-encoder RoBERTa-base variant without the entity-as-text input, so it is not a strict parameter-matched comparison against shared-encoder controls.

\subsection{Implementation}\label{implementation}

All variants use RoBERTa-base or RoBERTa-large encoders \citep{roberta}. The primary D2C-base Fusion system uses RoBERTa-base-family members; RoBERTa-large variants are analysis controls. Continuous evidence groups are projected through small MLPs before entering their pathways. Fusion weights are selected on development, the candidate pool is fixed before test evaluation, and no test-set interpolation tuning is used. The final fusion has three nonzero-weight members. Optimizer settings, fusion weights, scalar-detector proxy models, RACE rerun settings, and cost disclosures are provided in \hyperref[appendix-implementation]{Appendix~\ref*{appendix-implementation}}.

\section{Main Results}\label{main-results}

The main fine-grained analysis is four-way HH/HA/AH/AA low-FPR detection, because it tests whether a detector can distinguish off-diagonal mixed-origin cases rather than only detect any AI involvement. We first report the main four-way comparison, then use a compact control table to separate single-model method evidence from the development-selected detector system. Official Level-1/2/3 task collapses and scalar baseline details are reported in \hyperref[original-hart-protocol-references]{Appendix~\ref*{original-hart-protocol-references}}.

\subsection{Four-Way Low-FPR Results}\label{four-way-low-fpr-results}

\begin{table*}[!t]
\centering
\scriptsize
\setlength{\tabcolsep}{2.5pt}
\renewcommand{\arraystretch}{1.04}
\caption{Four-way low-FPR comparison. ``Published'' rows are representative RACE Table 2 values from a related HART reconstruction rather than exact-split reruns; ``MixD2C'' rows use the identical local split and evaluator. The dagger denotes our rerun of released RACE code. RACE reports Humanized before LLM-Generated; we remap these to AH and AA. The full published reference set and scalar official-task diagnostics are in \hyperref[mixd2c-split]{Appendix~\ref*{mixd2c-split}}--\hyperref[original-hart-protocol-references]{\ref*{original-hart-protocol-references}}.}
\label{tab:four-way-low-fpr}
\begin{adjustbox}{max width=\textwidth}
\begin{tabular*}{\textwidth}{@{\extracolsep{\fill}}llccccccc@{}}
\toprule
\multirow{2}{*}{\raisebox{-0.55ex}{Model}} & \multirow{2}{*}{\raisebox{-0.55ex}{Source}} & \multirow{2}{*}{\raisebox{-0.55ex}{AUROC}} & \multicolumn{4}{c}{TPR@1} & \multirow{2}{*}{\raisebox{-0.55ex}{Avg TPR@1}} & \multirow{2}{*}{\raisebox{-0.55ex}{F1-score}} \\
\cmidrule(lr){4-7}
 & & & HH & HA & AH & AA & & \\
\midrule
RoBERTa & Published & 0.9222 & 0.9936 & 0.6806 & 0.7092 & 0.6314 & 0.7537 & -- \\
RoBERTa-DANN & Published & 0.9617 & 0.9688 & 0.7503 & 0.7178 & 0.4889 & 0.7314 & -- \\
CoCo & Published & 0.9767 & 0.9968 & 0.7577 & 0.7943 & 0.6393 & 0.7970 & -- \\
LF-Motifs & Published & 0.9820 & 0.9668 & 0.6961 & 0.7562 & 0.6701 & 0.7723 & -- \\
RACE & Published & 0.9799 & 0.9904 & 0.8360 & 0.7541 & 0.7418 & 0.8306 & -- \\
\midrule
RACE-local\racedagger{} & MixD2C & 0.9849 & 0.9888 & 0.8463 & 0.7696 & 0.5752 & 0.7950 & 0.9030 \\
\textbf{D2C-Routing (3-seed)} & MixD2C & 0.9858 & 0.9871 & 0.8754 & 0.7435 & 0.7701 & 0.8440 & 0.9054 \\
\textbf{D2C-base Fusion} & MixD2C & \textbf{0.9889} & \textbf{0.9925} & \textbf{0.8888} & \textbf{0.7892} & \textbf{0.7708} & \textbf{0.8603} & \textbf{0.9195} \\
\bottomrule
\end{tabular*}
\end{adjustbox}
\end{table*}

Among the published references, RACE has the strongest Avg TPR@1\%FPR at 0.8306, but these values are contextual because exact published sample IDs are unavailable. On the exact MixD2C split, RACE-local is strong on ordinary metrics (0.9849 AUROC and 0.9030 F1-score), while its AA TPR@1\%FPR is 0.5752. D2C-Routing reaches 0.8440 Avg TPR@1\%FPR and 0.7701 on AA, but remains below RACE-local on AH (0.7435 versus 0.7696). The disclosed D2C-base Fusion system reaches 0.8603 Avg TPR@1\%FPR, with 0.7892 AH and 0.7708 AA. Thus the supported result is improved average and AA low-FPR ranking under the exact-split comparison, not uniform class-wise superiority or an exact reproduction of the published table.

\begin{figure}[!t]
\centering
\includegraphics[width=\columnwidth]{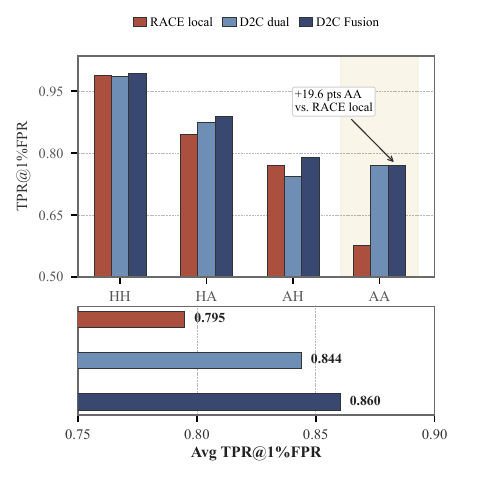}
\caption{Same-split low-FPR behavior on MixD2C. We compare RACE-local, D2C-Routing dual encoder, and D2C-base Fusion on the same MixD2C test split. Our rows mainly improve low-FPR ranking on AA, while RACE-local remains competitive on AH.}
\label{fig:main-low-fpr}
\end{figure}

\subsection{Ablations and Controls}\label{ablations-and-controls}

\begin{table*}[!t]
\centering
\scriptsize
\setlength{\tabcolsep}{2.2pt}
\renewcommand{\arraystretch}{1.04}
\setlength{\aboverulesep}{0.55ex}
\setlength{\belowrulesep}{0.55ex}
\caption{Matched controls and core ablations. Single-model rows use the same encoder family; cost rows match total parameter scale and disclose online forward calls. The D2C cost row reports a separately evaluated AH/AA-calibrated audit system; Table~\ref{tab:four-way-low-fpr} reports the final detector system.}
\label{tab:core-controls}
\begin{tabular*}{\textwidth}{@{\extracolsep{\fill}}>{\centering\arraybackslash}m{0.06\textwidth}>{\raggedright\arraybackslash}m{0.19\textwidth}>{\centering\arraybackslash}m{0.14\textwidth}>{\centering\arraybackslash}m{0.07\textwidth}>{\centering\arraybackslash}m{0.075\textwidth}>{\centering\arraybackslash}m{0.075\textwidth}>{\centering\arraybackslash}m{0.07\textwidth}>{\raggedright\arraybackslash}m{0.13\textwidth}@{}}
\toprule
\multicolumn{1}{c}{Group} & \multicolumn{1}{c}{Model} & \multicolumn{1}{c}{Setting} & \multicolumn{1}{c}{AUROC} & \multicolumn{1}{c}{Avg TPR@1} & \multicolumn{1}{c}{AA TPR@1} & \multicolumn{1}{c}{F1-score} & \multicolumn{1}{c}{Role} \\
\midrule
\multirow[c]{4}{*}{Single} & RoBERTa-base & 3 seeds & 0.9832 & 0.7666 & 0.5447 & 0.8940 & Text-only \\
 & Flat concat & 3 seeds & 0.9851 & 0.7923 & 0.6113 & 0.8987 & Flat features \\
 & \textbf{D2C-Routing} & shared, 3 seeds & 0.9846 & 0.8110 & 0.6841 & 0.8995 & Shared encoder \\
 & \textbf{D2C-Routing} & dual, 3 seeds & 0.9858 & 0.8440 & 0.7701 & 0.9054 & Best single model \\
\cmidrule(lr){2-8}
\multirow[c]{3}{*}{Cost} & Text-only 5$\times$ uniform & 623.24M / 5 calls & 0.9859 & 0.8241 & 0.7385 & 0.8932 & Cost control \\
 & Flat features 5$\times$ uniform & 623.87M / 5 calls & 0.9880 & 0.8443 & \textbf{0.7772} & 0.9093 & Cost control \\
 & \textbf{D2C cost-audit system} & 625.34M / 3 calls & \textbf{0.9894} & \textbf{0.8601} & 0.7686 & \textbf{0.9185} & Calibrated system \\
\cmidrule(lr){2-8}
\multirow[c]{2}{*}{\raisebox{-1.4ex}{Ablation}} & No dimension loss & 3 seeds & 0.9834 & 0.7801 & 0.6662 & 0.8994 & No dimension supervision \\
 & No gate & 3 seeds & 0.9818 & 0.7838 & 0.6870 & 0.8795 & No learned composition \\
\bottomrule
\end{tabular*}
\end{table*}

\begin{table*}[!t]
\centering
\scriptsize
\setlength{\tabcolsep}{2.0pt}
\renewcommand{\arraystretch}{1.06}
\caption{Supporting diagnostics for transfer, mechanism, and error structure. Main four-way results are reported in \hyperref[tab:four-way-low-fpr]{Table~\ref*{tab:four-way-low-fpr}}.}
\label{tab:supporting-diagnostics}
\makebox[\textwidth][c]{%
\begin{tabular}{@{}>{\raggedright\arraybackslash}m{0.185\textwidth}>{\raggedright\arraybackslash}m{0.195\textwidth}>{\raggedright\arraybackslash}m{0.345\textwidth}>{\raggedright\arraybackslash}m{0.235\textwidth}@{}}
\toprule
Diagnostic & Mode & Result & Interpretation \\
\midrule
Dimension heads & Source heads & C 0.9937; E 0.9871 AUROC & Targets learnable \\
Composition controls & Learned vs fixed/concat/hard & Avg 0.8110 vs 0.7754/0.7942/0.7868 & Learned composition helps \\
Grouped robustness & Matched systems & F1 0.9100; Avg 0.8507 & Competitive, not uniformly significant \\
External expression & APT / PAN zero-shot & AUROC 0.8993 / 0.9392 & Dimension-targeted transfer \\
External content & PAN zero-shot & AUROC 0.8012 vs text 0.7961 & Small controlled gain \\
AH diagnosis & Branch accuracy & Content 0.9477; expression 0.6438 & Expression bottleneck \\
\bottomrule
\end{tabular}%
}
\end{table*}

The controls separate architecture evidence from detector-system engineering. Shared D2C improves low-FPR ranking over same-family text-only and flat controls, while the dual encoder is the strongest single model. At closely matched total scale, the D2C cost-audit system improves Macro-F1 and AUROC over both five-member controls and Avg TPR@1 over text-only. Its Avg TPR advantage over flat 5$\times$ is positive but not significant at 95\%, and flat 5$\times$ has higher AA TPR. Removing dimension supervision or learned composition weakens four-way low-FPR performance. Full feature removals, paired intervals, routing controls, and composition alternatives are provided in \hyperref[additional-architecture-controls]{Appendix~\ref*{additional-architecture-controls}} and \hyperref[composition-controls]{Appendix~\ref*{composition-controls}}.

\subsection{Diagnostics}\label{diagnostics}

\hyperref[tab:supporting-diagnostics]{Table~\ref*{tab:supporting-diagnostics}} summarizes mechanism, grouped-split, transfer, and error diagnostics. Group-aware HART evaluates a stricter in-benchmark split; APT-Eval and PAN evaluate zero-shot expression- or content-origin scores; FAIDSet and the HART-style MixSet reconstruction evaluate supervised external adaptation. Expression-origin transfer is the clearest external result, whereas PAN content transfer is small relative to a matched text-only score and direct four-way transfer to MixSet is negative. The AH branch diagnosis further localizes the single-model weakness to expression-origin recognition. Full protocols and negative results are in \hyperref[transfer-checks]{Appendix~\ref*{transfer-checks}}--\hyperref[mechanism-diagnostics]{\ref*{mechanism-diagnostics}}.

\FloatBarrier
\section{Conclusion}\label{conclusion}

We presented D2C-Routing for mixed-origin AI-text detection, with supervised content-origin and expression-origin prediction followed by learned composition. On MixD2C, D2C-Routing improves average and AA low-FPR ranking over same-split single-model controls and RACE-local, while AH remains the hardest boundary. D2C-base Fusion provides the strongest disclosed detector-system result, and cost-matched controls show that its ordinary-metric gains are not explained by ensemble scale alone. Across targeted transfer evaluations, expression-origin signals transfer more reliably than full four-way predictions. Together, the results support supervised source dimensions and learned composition while identifying AH expression recognition as the main remaining challenge.

\section*{Limitations}\label{limitations}

The main positive result is in-domain on MixD2C. Direct HART-to-MixSet transfer reaches 0.2262 Macro-F1 and 0.4372 Macro AUROC, so the external experiments support dimension-specific transfer and supervised adaptation rather than broad out-of-domain four-way generalization.

We run released RACE code on the same MixD2C split, but published RACE values remain related-reconstruction references because frozen sample IDs and checkpoints are unavailable. D2C-base Fusion is a development-selected detector system, whereas the single-model D2C-Routing results provide the architecture-level evidence. AH remains the main single-model weakness: content-origin accuracy is 0.9477 but expression-origin accuracy is 0.6438 on AH.

ModernBERT does not yield a reliable D2C-over-text advantage, and correct, swapped, and fixed-random routing are statistically similar on the principal low-FPR metric. The current evidence therefore supports dimension supervision and learned composition, but it does not establish a uniquely optimal or causally interpretable handcrafted feature assignment.

\section*{Ethical Considerations}\label{ethical-considerations}

This work studies detection of AI-generated and mixed-origin writing. Such detectors should not be used as sole evidence for punitive decisions about authorship, academic integrity, employment, or access to services. The labels in MixD2C describe controlled construction protocols rather than a person's intent or honesty, and low-FPR evaluation does not remove the risk of false accusations. The intended use is therefore decision support, auditing, and research on mixed-origin text, with human review and context-specific policy safeguards. The data are derived from public research benchmarks, but benchmark distributions may underrepresent writers, domains, languages, and editing practices outside the evaluated setting.

\section*{Acknowledgments}

This research work is supported by the National Key Research and Development Program of China under Grant NO. 2024YFF0729003, the National Natural Science Foundation of China under Grant NOs. 62176014, the Fundamental Research Funds for the Central Universities.

\bibliography{references}

\clearpage
\appendix
\setcounter{table}{0}
\renewcommand{\thetable}{A\arabic{table}}
\section{Additional Results and Baselines}\label{additional-results-and-baselines}

\subsection{MixD2C Split}\label{mixd2c-split}

MixD2C merges the released HART development and test files, then stratifies by domain and RACE-style class into the following train/dev/test split. AH is the minority class in this reconstruction.

\begin{table*}[!tbp]
\centering
\scriptsize
\setlength{\tabcolsep}{2.5pt}
\renewcommand{\arraystretch}{1.00}
\caption{MixD2C split. Released HART development and test files are merged and split by domain and RACE-style class.}
\label{tab:mixd2c-split-counts}
\begin{adjustbox}{max width=0.48\textwidth}
\begin{tabular*}{0.48\textwidth}{@{\extracolsep{\fill}}lccccc@{}}
\toprule
Split & HH & HA & AH & AA & Total \\
\midrule
Train & 2,800 & 2,800 & 710 & 4,890 & 11,200 \\
Dev & 400 & 400 & 101 & 699 & 1,600 \\
Test & 800 & 800 & 204 & 1,396 & 3,200 \\
\bottomrule
\end{tabular*}
\end{adjustbox}
\end{table*}

RACE Table 2 is the closest published four-way reference for HH/HA/AH/AA comparison. The rows below are external references under the same RACE-style HART construction, not local reruns from frozen sample IDs in this repository. RACE reports class columns in the order Human-Written, LLM-Polished, LLM-Generated, Humanized. We map them to HH, HA, AA, AH, then present them in HH, HA, AH, AA order; therefore the Humanized and LLM-Generated columns are swapped in our display.

\begin{table*}[!tbp]
\centering
\scriptsize
\setlength{\tabcolsep}{2.5pt}
\renewcommand{\arraystretch}{1.00}
\caption{Full RACE Table 2 reference set, displayed in HH/HA/AH/AA order.}
\label{tab:race-full-reference}
\begin{tabular*}{\textwidth}{@{\extracolsep{\fill}}lllllll@{}}
\toprule
Method & Macro AUROC & HH TPR@1 & HA TPR@1 & AH TPR@1 & AA TPR@1 & Avg TPR@1 \\
\midrule
RoBERTa & 0.9222 & 0.9936 & 0.6806 & 0.7092 & 0.6314 & 0.7537 \\
RoBERTa-DANN & 0.9617 & 0.9688 & 0.7503 & 0.7178 & 0.4889 & 0.7314 \\
CoCo & 0.9767 & 0.9968 & 0.7577 & 0.7943 & 0.6393 & 0.7970 \\
SeqXGPT & 0.8987 & 0.9838 & 0.1523 & 0.3168 & 0.1432 & 0.3990 \\
DeTeCtive & 0.9574 & 0.9862 & 0.0000 & 0.7723 & 0.0000 & 0.4396 \\
LF-Motifs & 0.9820 & 0.9668 & 0.6961 & 0.7562 & 0.6701 & 0.7723 \\
Binoculars-MLP & 0.7915 & 0.2949 & 0.0734 & 0.0550 & 0.0437 & 0.1170 \\
Binoculars-C-T & 0.5003 & 0.0000 & 0.0000 & 0.0000 & 0.0000 & 0.0000 \\
Fast-DetectGPT & 0.6170 & 0.0000 & 0.0337 & 0.0009 & 0.2627 & 0.0770 \\
Fast-DetectGPT-MLP & 0.7369 & 0.0312 & 0.0387 & 0.0396 & 0.2935 & 0.1080 \\
Fast-DetectGPT-C-T & 0.4993 & 0.0000 & 0.0000 & 0.0000 & 0.0000 & 0.0000 \\
TDT-SVC & 0.5716 & 0.0288 & 0.0237 & 0.0050 & 0.0358 & 0.0233 \\
RACE & 0.9799 $\pm$ 0.0013 & 0.9904 $\pm$ 0.0040 & 0.8360 $\pm$ 0.0161 & 0.7541 $\pm$ 0.0103 & 0.7418 $\pm$ 0.0095 & 0.8306 $\pm$ 0.0057 \\
\bottomrule
\end{tabular*}
\end{table*}

The local rerun below uses the identical MixD2C split used by our models. It is not a reproduction of the published RACE table because frozen published sample IDs and checkpoints are not packaged with the released repository.

\begin{table*}[!tbp]
\centering
\scriptsize
\setlength{\tabcolsep}{2.5pt}
\renewcommand{\arraystretch}{1.00}
\caption{Local public-code RACE rerun on the exact MixD2C split.}
\label{tab:race-local-rerun}
\begin{tabular*}{\textwidth}{@{\extracolsep{\fill}}llllllll@{}}
\toprule
Local run & Macro AUROC & Macro-F1 & HH TPR@1 & HA TPR@1 & AH TPR@1 & AA TPR@1 & Avg TPR@1 \\
\midrule
\textbf{RACE-local, 20ep strict rstdt} & 0.9849 & 0.9030 & 0.9888 & 0.8463 & 0.7696 & 0.5752 & 0.7950 \\
\bottomrule
\end{tabular*}
\end{table*}

\subsection{Original HART Protocol References}\label{original-hart-protocol-references}

HART's original detector tables are not directly comparable to MixD2C. They use the original HART protocol and mostly report official binary risk tasks rather than four-way low-FPR diagnostics. We therefore use them as benchmark context, not as same-split baselines. The full English table from HART is reproduced below to keep the source of the original-protocol references auditable.

\begin{table*}[!tbp]
\centering
\scriptsize
\setlength{\tabcolsep}{2.5pt}
\renewcommand{\arraystretch}{1.00}
\caption{Original HART English detector references under the original HART protocol.}
\label{tab:hart-original-reference}
\resizebox{\textwidth}{!}{%
\begin{tabular}{lllllllllllll}
\toprule
Detector & L3 Essay & L3 ArXiv & L3 Writing & L3 ALL & L2 Essay & L2 ArXiv & L2 Writing & L2 ALL & L1 Essay & L1 ArXiv & L1 Writing & L1 ALL \\
\midrule
RoBERTa(ChatGPT) & 0.636 & 0.796 & 0.653 & 0.662 (16\%) & 0.435 & 0.687 & 0.498 & 0.502 (8\%) & 0.471 & 0.955 & 0.606 & 0.566 (9\%) \\
RADAR & 0.692 & 0.849 & 0.647 & 0.728 (14\%) & 0.566 & 0.814 & 0.630 & 0.687 (10\%) & 0.705 & 0.857 & 0.700 & 0.758 (20\%) \\
Log-Perplexity & 0.868 & 0.850 & 0.811 & 0.799 (33\%) & 0.364 & 0.485 & 0.438 & 0.473 (11\%) & 0.769 & 0.530 & 0.625 & 0.576 (6\%) \\
Log-Rank & 0.867 & 0.874 & 0.813 & 0.814 (39\%) & 0.380 & 0.460 & 0.441 & 0.465 (11\%) & 0.739 & 0.542 & 0.611 & 0.573 (8\%) \\
LRR & 0.835 & 0.909 & 0.797 & 0.840 (50\%) & 0.560 & 0.616 & 0.551 & 0.573 (25\%) & 0.616 & 0.576 & 0.558 & 0.568 (19\%) \\
Glimpse & 0.929 & 0.869 & 0.819 & 0.849 (58\%) & 0.754 & 0.737 & 0.625 & 0.676 (30\%) & 0.878 & 0.719 & 0.618 & 0.688 (22\%) \\
Fast-Detect & 0.883 & 0.877 & 0.840 & 0.862 (60\%) & 0.734 & 0.718 & 0.692 & 0.711 (47\%) & 0.877 & 0.769 & 0.740 & 0.778 (55\%) \\
C2 (Fast-Detect) & 0.734 & 0.787 & 0.765 & 0.738 (18\%) & 0.778 & 0.862 & 0.819 & 0.798 (42\%) & 0.712 & 0.779 & 0.742 & 0.730 (33\%) \\
C2-T (Fast-Detect) & 0.864 & 0.896 & 0.890 & 0.876 (61\%) & 0.785 & 0.915 & 0.890 & 0.855 (59\%) & 0.907 & 0.849 & 0.836 & 0.843 (59\%) \\
Binoculars & 0.897 & 0.882 & 0.847 & 0.870 (62\%) & 0.735 & 0.715 & 0.693 & 0.711 (44\%) & 0.879 & 0.769 & 0.740 & 0.780 (55\%) \\
C2 (Binoculars) & 0.736 & 0.789 & 0.770 & 0.737 (17\%) & 0.781 & 0.856 & 0.822 & 0.791 (35\%) & 0.701 & 0.761 & 0.743 & 0.716 (25\%) \\
C2-T (Binoculars) & 0.854 & 0.904 & 0.905 & 0.883 (61\%) & 0.746 & 0.913 & 0.895 & 0.848 (32\%) & 0.900 & 0.840 & 0.828 & 0.838 (58\%) \\
\bottomrule
\end{tabular}
}
\end{table*}

In the English ALL setting, HART reports C2-T Fast-Detect as the strongest Level-1 and Level-2 reference among its listed original-protocol detectors, with AUROC 0.843 and 0.855 and TPR@5\%FPR 59\% for both tasks; for Level-3, C2-T Binoculars reaches AUROC 0.883 and TPR@5\%FPR 61\%. These numbers support the need to report Level-1/2/3, but they remain original-protocol context rather than same-split MixD2C baselines.

The table below reports the local MixD2C scalar-detector reruns used to contextualize the official Level-1/2/3 task collapses. Binoculars is included as a proxy implementation because the official Falcon-7B/Falcon-7B-Instruct pair was not used on the 12GB local GPU.

\begin{table*}[!tbp]
\centering
\scriptsize
\setlength{\tabcolsep}{2.5pt}
\renewcommand{\arraystretch}{1.00}
\caption{Local MixD2C scalar-detector reruns for official Level-1/2/3 collapses.}
\label{tab:scalar-reruns}
\begin{tabular*}{\textwidth}{@{\extracolsep{\fill}}lllll@{}}
\toprule
Detector & Proxy LM / model & L1 AUROC / F1 / TPR@5 & L2 AUROC / F1 / TPR@5 & L3 AUROC / F1 / TPR@5 \\
\midrule
SpecDetect & gpt2 & 0.6859 / 0.8808 / 0.2137 & 0.5877 / 0.6862 / 0.1744 & 0.6098 / 0.6259 / 0.2013 \\
SpecDetect & gpt2-xl & 0.7129 / 0.8864 / 0.2892 & 0.6415 / 0.6937 / 0.2694 & 0.6693 / 0.6333 / 0.2958 \\
Fast-DetectGPT analytic & gpt2-xl scoring/reference & 0.7205 / 0.8582 / 0.3688 & 0.6581 / 0.6676 / 0.3419 & 0.6883 / 0.6179 / 0.3725 \\
Binoculars proxy & observer=gpt2-xl, performer=gpt2 & 0.6314 / 0.8571 / 0.2687 & 0.5588 / 0.6667 / 0.2256 & 0.5845 / 0.6075 / 0.2357 \\
\bottomrule
\end{tabular*}
\end{table*}

The next table reports same-split official collapses for RACE-local and D2C-base Fusion. Each cell reports AUROC / F1 / TPR@5\%FPR. These MixD2C values should not be compared as exact head-to-head results against HART's original-protocol detector table above.

\begin{table*}[!tbp]
\centering
\scriptsize
\setlength{\tabcolsep}{2.5pt}
\renewcommand{\arraystretch}{1.00}
\caption{Same-split official-task collapses on MixD2C. Each cell reports AUROC / F1 / TPR@5\%FPR.}
\label{tab:additional-8f4e38ab}
\begin{tabular*}{\textwidth}{@{\extracolsep{\fill}}lllll@{}}
\toprule
Model & Source & Level-1 & Level-2 & Level-3 \\
\midrule
SpecDetect-gpt2-xl & local scalar rerun & 0.7129 / 0.8864 / 0.2892 & 0.6415 / 0.6937 / 0.2694 & 0.6693 / 0.6333 / 0.2958 \\
Fast-DetectGPT analytic & local scalar rerun & 0.7205 / 0.8582 / 0.3688 & 0.6581 / 0.6676 / 0.3419 & 0.6883 / 0.6179 / 0.3725 \\
RACE-local & same-split local rerun & 0.9985 / 0.9925 / 0.9967 & 0.9943 / 0.9641 / 0.9731 & 0.9841 / 0.9365 / 0.9398 \\
\textbf{D2C-base Fusion} & ours, probability collapse & \textbf{0.9988 / 0.9925 / 0.9962} & \textbf{0.9959 / 0.9702 / 0.9850} & \textbf{0.9906 / 0.9498 / 0.9685} \\
\bottomrule
\end{tabular*}
\end{table*}

\subsection{Adjacent Mixed-Origin Work}\label{adjacent-mixed-origin-work}

Several recent datasets and benchmarks study mixed or collaborative authorship but are not numeric baselines for \hyperref[tab:four-way-low-fpr]{Table~\ref*{tab:four-way-low-fpr}}. APT-Eval focuses on AI-polished writing and detector reliability \citep{saha2025almost}; M4, M4GT-Bench, HC3, and RAID provide broader generated-text detection benchmarks \citep{guo2023hc3,wang-etal-2024-m4,wang2024m4gtbench,dugan2024raid}; RoFT evaluates localized human-machine boundary detection \citep{roft-boundary}; RealBench/DETree studies manipulated or hierarchical collaborative processes \citep{wang-etal-2025-real,detree}. These lines of work motivate the problem and should be discussed as related settings, while HART and RACE remain the directly comparable HART-family sources for the present four-way evaluation.

Truth Mirror and RAID-style files expose binary, content, or task labels rather than the full HH/HA/AH/AA content-expression composition used here. We therefore use them for transfer or robustness evaluation rather than as main four-way baselines.

\subsection{Additional Architecture Controls}\label{additional-architecture-controls}

The following controls use MixD2C and the unified evaluator. The first table contains seed-42 architecture checks; the subsequent three-seed tables test modern backbones and routing alignment. Together, they evaluate architecture sensitivity independently of final detector-system selection.

\begin{table*}[!tbp]
\centering
\scriptsize
\setlength{\tabcolsep}{2.5pt}
\renewcommand{\arraystretch}{1.00}
\caption{Supplemental architecture controls on MixD2C.}
\label{tab:architecture-controls}
\resizebox{\textwidth}{!}{%
\begin{tabular}{lllllllll}
\toprule
Model & Macro AUROC & Avg TPR@1 & Macro-F1 & HH TPR@1 & HA TPR@1 & AH TPR@1 & AA TPR@1 & Use in paper \\
\midrule
\textbf{D2C-Routing (base no-syntax + AA rank)} & 0.9845 & 0.8206 & 0.9051 & 0.9888 & 0.8613 & 0.7549 & 0.6777 & Routed baseline \\
Frozen \textbf{D2C-Routing} (last layer) & 0.9581 & 0.5444 & 0.8165 & 0.9363 & 0.2825 & 0.5735 & 0.3854 & Frozen-probe control \\
Flat all-features RoBERTa-base & 0.9873 & 0.7982 & 0.9071 & 0.9875 & 0.8138 & 0.7549 & 0.6368 & Feature-stacking control \\
\textbf{D2C-Routing (DistilRoBERTa)} & 0.9776 & 0.7414 & 0.8691 & 0.9788 & 0.7425 & 0.6912 & 0.5530 & Small-backbone control \\
\textbf{D2C-Routing (Longformer-base, 512 tokens)} & 0.9802 & 0.7639 & 0.8815 & 0.9825 & 0.7413 & 0.7059 & 0.6261 & Backbone/length control \\
\textbf{D2C-Routing (DeBERTa-base no-syntax)} & 0.9867 & 0.8158 & 0.9005 & 0.9875 & 0.8338 & 0.7451 & 0.6970 & Alternative backbone control \\
\bottomrule
\end{tabular}
}
\end{table*}

The frozen-last1 control shows that D2C-Routing is not merely a frozen encoder probe: Avg TPR@1\%FPR drops from 0.8206 to 0.5444, and AA TPR@1\%FPR drops from 0.6777 to 0.3854. The flat all-features control is competitive on ordinary Macro AUROC and Macro-F1, but lower on low-FPR ranking, especially AA. DistilRoBERTa and Longformer-base at 512 tokens both underperform the routed RoBERTa-base control. DeBERTa-base improves the single-backbone low-FPR result over the RoBERTa-base seed42 routed baseline, especially for AA, but still remains below the final D2C-base Fusion system.

\begin{table*}[!tbp]
\centering
\scriptsize
\setlength{\tabcolsep}{3pt}
\renewcommand{\arraystretch}{1.00}
\caption{Three-seed modern-backbone comparisons on MixD2C. Values are mean $\pm$ standard deviation.}
\label{tab:modern-backbones}
\begin{tabular*}{\textwidth}{@{\extracolsep{\fill}}lrrrrrr@{}}
\toprule
Model & Params & Macro AUROC & Macro-F1 & Avg TPR@1 & AH TPR@1 & AA TPR@1 \\
\midrule
ModernBERT text-only & 149.02M & $0.9850\pm0.0022$ & $0.9089\pm0.0076$ & $0.8032\pm0.0136$ & $0.7484\pm0.0172$ & $0.6562\pm0.0389$ \\
ModernBERT D2C & 149.73M & $0.9851\pm0.0011$ & $0.9026\pm0.0025$ & $0.8073\pm0.0138$ & $0.7435\pm0.0113$ & $0.6796\pm0.0366$ \\
DeBERTa-base D2C & -- & $0.9877\pm0.0009$ & $0.9094\pm0.0076$ & $0.8320\pm0.0157$ & $0.7614\pm0.0142$ & $0.7051\pm0.0372$ \\
\bottomrule
\end{tabular*}
\end{table*}

\begin{table*}[!tbp]
\centering
\scriptsize
\setlength{\tabcolsep}{3pt}
\renewcommand{\arraystretch}{1.00}
\caption{Three-seed comparison of correct, swapped, fixed-random, and flat evidence routing on MixD2C.}
\label{tab:routing-alignment}
\begin{tabular*}{\textwidth}{@{\extracolsep{\fill}}lrrrrr@{}}
\toprule
Routing & Macro AUROC & Macro-F1 & Avg TPR@1 & AH TPR@1 & AA TPR@1 \\
\midrule
Correct role-aligned & $0.9846\pm0.0010$ & $0.8995\pm0.0100$ & $0.8110\pm0.0165$ & $0.7435\pm0.0242$ & $0.6841\pm0.0310$ \\
Swapped & $0.9845\pm0.0033$ & $0.8950\pm0.0157$ & $0.8076\pm0.0254$ & $0.7222\pm0.0396$ & $0.6991\pm0.0326$ \\
Fixed random & $0.9849\pm0.0005$ & $0.9058\pm0.0057$ & $0.8034\pm0.0109$ & $0.7451\pm0.0147$ & $0.6476\pm0.0591$ \\
Flat concat & $0.9851\pm0.0019$ & $0.8987\pm0.0079$ & $0.7923\pm0.0072$ & $0.7516\pm0.0057$ & $0.6113\pm0.0499$ \\
\bottomrule
\end{tabular*}
\end{table*}

These controls do not prove that the handcrafted feature-to-dimension assignment is uniquely optimal. The supported mechanism is the use of supervised source dimensions with learned composition; the exact routing partition remains an inductive bias.

\subsection{Transfer Checks}\label{transfer-checks}

The following analyses evaluate transfer when the domain or dataset changes.

\textbf{Group-aware HART split.} To test whether the model relies heavily on variant or topic overlap, we evaluate a stricter split in which all variants of the same base text stay in the same partition. The split contains 4000 groups with zero cross-split leaks.

\begin{table*}[!tbp]
\centering
\scriptsize
\setlength{\tabcolsep}{2.5pt}
\renewcommand{\arraystretch}{1.00}
\caption{Group-aware MixD2C results with matched detector systems. Paired intervals for D2C versus text-only cross zero for Macro-F1 and Avg TPR@1; only the accuracy difference is significant.}
\label{tab:group-aware}
\begin{tabular*}{\textwidth}{@{\extracolsep{\fill}}lrrrl@{}}
\toprule
Model/system & Macro-F1 & Macro AUROC & Avg TPR@1 & Interpretation \\
\midrule
Text-only dev-selected ensemble & 0.9047 & \textbf{0.9883} & 0.8289 & Matched ensemble \\
Flat-concat dev-selected ensemble & 0.8903 & 0.9863 & 0.8191 & Matched ensemble \\
\textbf{Calibrated D2C system} & \textbf{0.9100} & 0.9877 & \textbf{0.8507} & Competitive grouped robustness \\
\bottomrule
\end{tabular*}
\end{table*}

Performance drops under grouped splitting, so this is a robustness sanity check rather than evidence of domain-invariant detection. The calibrated D2C system remains competitive with matched text-only and flat ensembles; its accuracy advantage over text-only is significant, whereas Macro-F1 and Avg TPR@1 intervals cross zero.

\textbf{Leave-one-domain-out HART.} Each row trains on the other three HART domains and tests on the held-out domain.

\begin{table*}[!tbp]
\centering
\scriptsize
\setlength{\tabcolsep}{2.5pt}
\renewcommand{\arraystretch}{1.00}
\caption{Leave-one-domain-out HART transfer checks.}
\label{tab:lodo-transfer}
\begin{tabular*}{\textwidth}{@{\extracolsep{\fill}}llllll@{}}
\toprule
Held-out domain & Model & Macro AUROC & Avg TPR@1 & Macro-F1 & AA TPR@1 \\
\midrule
Essay & \textbf{D2C-Routing} & 0.9000 & 0.3663 & 0.6414 & 0.1834 \\
Essay & Text-only & 0.8894 & 0.3094 & 0.6198 & 0.0000 \\
ArXiv & \textbf{D2C-Routing} & 0.9495 & 0.6760 & 0.6935 & 0.8006 \\
ArXiv & Text-only & 0.9641 & 0.7464 & 0.8051 & 0.7721 \\
News & \textbf{D2C-Routing} & 0.9096 & 0.3847 & 0.5555 & 0.3305 \\
News & Text-only & 0.9124 & 0.4363 & 0.5827 & 0.3675 \\
Writing & \textbf{D2C-Routing} & 0.8884 & 0.3062 & 0.4622 & 0.4087 \\
Writing & Text-only & 0.8834 & 0.3088 & 0.5315 & 0.4696 \\
\bottomrule
\end{tabular*}
\end{table*}

\textbf{RAID binary transfer.} RAID provides binary content or task labels rather than HH/HA/AH/AA labels, so we use it to evaluate content-origin transfer.

\begin{table*}[!tbp]
\centering
\scriptsize
\setlength{\tabcolsep}{2.5pt}
\renewcommand{\arraystretch}{1.00}
\caption{RAID binary transfer checks.}
\label{tab:raid-transfer}
\begin{tabular*}{\textwidth}{@{\extracolsep{\fill}}lllllll@{}}
\toprule
Dataset & Model & AUROC & F1 & Accuracy & TPR@1 & TPR@5 \\
\midrule
RAID test & Text-only & 0.7312 & 0.5464 & 0.6493 & 0.2650 & 0.3435 \\
RAID test & \textbf{D2C-Routing (no-external)} & 0.7210 & 0.5713 & 0.6270 & 0.2010 & 0.2550 \\
RAID nonnative & Text-only & 0.9349 & 0.8639 & 0.8736 & 0.5495 & 0.7363 \\
RAID nonnative & \textbf{D2C-Routing (no-external)} & 0.9144 & 0.8095 & 0.8242 & 0.2198 & 0.6374 \\
\bottomrule
\end{tabular*}
\end{table*}

\textbf{Zero-shot polished-writing transfer.} We also evaluate a 3-seed equal-weight D2C-Routing (no-external) ensemble on APT-Eval without retraining and without external-target development tuning. This setting removes HART-side pre-extracted RST, entity, and style features because APT-Eval does not provide the same feature cache.

\begin{table*}[!tbp]
\centering
\scriptsize
\setlength{\tabcolsep}{2.5pt}
\renewcommand{\arraystretch}{1.00}
\caption{Zero-shot APT-Eval expression-origin transfer.}
\label{tab:additional-b7c0fcb5}
\begin{adjustbox}{max width=\textwidth}
\begin{tabular}{llllllll}
\toprule
Dataset / setting & Score or mode & N & AUROC & F1 / Macro-F1 & TPR@1 & TPR@5 & Note \\
\midrule
\textbf{APT-Eval paired original vs polished} & expression-head AI-expression prob & 300 pairs & - & - & - & - & Polished version ranked higher in \textbf{97.7\%} of pairs \\
\textbf{APT-Eval HH vs HA} & expression-head AI-expression prob & 14,950 & \textbf{0.8993} & \textbf{0.7907} & \textbf{0.3658} & \textbf{0.6428} & Human original vs AI-polished human text \\
APT-Eval HH vs HA & content-head AI-content prob & 14,950 & 0.5596 & 0.3011 & 0.0831 & 0.1843 & Dimension-control score \\
\bottomrule
\end{tabular}
\end{adjustbox}
\end{table*}

\begin{table*}[!tbp]
\centering
\scriptsize
\setlength{\tabcolsep}{2.5pt}
\renewcommand{\arraystretch}{1.00}
\caption{Protocol-separated robustness evidence. FAIDSet and the HART-style MixSet reconstruction use target supervision; APT/PAN use HART-trained scores without target tuning. MixSet four-way labels are reconstructed from released transformation provenance, not native benchmark labels.}
\label{tab:protocol-separated-transfer}
\begin{tabularx}{\textwidth}{@{}p{0.20\textwidth}p{0.19\textwidth}p{0.29\textwidth}X@{}}
\toprule
Setting & Supervision & Result & Claim role \\
\midrule
FAIDSet clean test \citep{ta-etal-2026-faid} & Target-supervised adaptation & Accuracy 0.9618; Macro-F1 0.9599 & External adaptation, current public split \\
APT-Eval HH/HA \citep{saha2025almost} & Zero-shot expression score & AUROC 0.8993; 293/300 paired ranking & Expression-origin transfer \\
PAN LLM-DetectAIve HH/HA \citep{pan2025generated} & Zero-shot expression score & AUROC 0.9392; TPR@1 0.6491 ($N=18{,}480$) & Expression-origin transfer \\
PAN machine-mediated content contrast & Zero-shot content score & AUROC 0.8012 vs matched text 0.7961 ($N=20{,}703$) & Small content-origin transfer \\
MixSet HART-style reconstruction \citep{zhang-etal-2024-mixset} & Target-supervised adaptation & Accuracy 0.8504; Macro-F1 0.7811 & External reconstructed four-process task \\
MixSet direct four-way transfer & Zero-shot, no adaptation & Macro-F1 0.2262; AUROC 0.4372 & Negative broad-OOD result \\
\bottomrule
\end{tabularx}
\end{table*}

Because the protocols differ, we interpret these results separately. APT/PAN evaluate targeted dimension transfer, while FAIDSet and MixSet measure supervised adaptation under different label spaces and construction protocols. On the controlled PAN content contrast, the D2C content score has only a small AUROC advantage over matched text-only and no consistent low-FPR advantage. Direct MixSet transfer remains weak, leaving broad four-way zero-shot generalization unresolved.

\subsection{Direct Evidence Probes}\label{direct-evidence-probes}

Before testing full neural systems, we ask whether the routed evidence groups contain source-origin signal on their own. We train simple logistic probes using pre-extracted evidence features. These probes measure association between evidence groups and source dimensions.

\begin{table*}[!tbp]
\centering
\scriptsize
\setlength{\tabcolsep}{2.5pt}
\renewcommand{\arraystretch}{1.00}
\caption{Direct evidence probe results. The table reports dimension-level and four-way probe performance from pre-extracted evidence features.}
\label{tab:d2c-ablations}
\begin{tabular*}{\textwidth}{@{\extracolsep{\fill}}lllll@{}}
\toprule
Feature group & Content-origin AUROC & Expression-origin AUROC & Four-way Macro AUROC & Four-way Macro-F1 \\
\midrule
Entity/coherence & 0.8261 & 0.7277 & 0.7460 & 0.4363 \\
Expression features & 0.9300 & 0.8515 & 0.8777 & 0.5907 \\
RST discourse & 0.8240 & 0.7250 & 0.7242 & 0.4100 \\
Entity + RST & 0.8842 & 0.7747 & 0.7843 & 0.4794 \\
All evidence & 0.9408 & 0.8595 & 0.8781 & 0.6150 \\
\bottomrule
\end{tabular*}
\end{table*}

\begin{figure}[!htbp]
\centering
\includegraphics[width=\columnwidth]{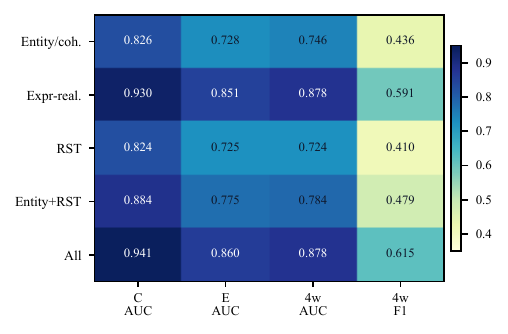}
\caption{Heatmap summary of source-dimension and four-way evidence probes.}
\label{fig:evidence-probes}
\end{figure}

Entity/coherence and RST features carry substantial content-origin signal, and their combination reaches 0.8842 content-origin AUROC. The expression feature block reaches 0.8515 expression-origin AUROC, but it is also strongly predictive of content origin. This cross-dimension signal motivates explicit dimension supervision and learned composition while cautioning against rigid feature-role interpretations.

\subsection{Composition Controls}\label{composition-controls}

These tables provide the full feature, expression-subgroup, and composition controls summarized in the main text. All rows are 3-seed means under the RoBERTa-base no-syntax + AA-ranking setup unless stated otherwise.

\begin{table}[!tbp]
\centering
\scriptsize
\setlength{\tabcolsep}{2pt}
\renewcommand{\arraystretch}{1.00}
\caption{Feature-removal ablations.}
\label{tab:feature-removal}
\resizebox{\columnwidth}{!}{%
\begin{tabular}{lrrrr}
\toprule
Model & AUROC & Avg TPR@1 & Macro-F1 & AA TPR@1 \\
\midrule
\textbf{D2C-Routing (full)} & 0.9846 & 0.8110 & 0.8995 & 0.6841 \\
No RST & 0.9814 & 0.7639 & 0.8869 & 0.5447 \\
No entity/coherence & 0.9833 & 0.7868 & 0.8969 & 0.6098 \\
No expression features & 0.9802 & 0.7541 & 0.8831 & 0.5805 \\
No routed external features & 0.9843 & 0.7988 & 0.8948 & 0.6655 \\
\bottomrule
\end{tabular}%
}
\end{table}

\begin{table}[!tbp]
\centering
\scriptsize
\setlength{\tabcolsep}{2pt}
\renewcommand{\arraystretch}{1.00}
\caption{Expression-side subgroup ablations.}
\label{tab:expression-subgroups}
\resizebox{\columnwidth}{!}{%
\begin{tabular}{lrrrr}
\toprule
Setting & AUROC & Avg TPR@1 & Macro-F1 & AA TPR@1 \\
\midrule
\textbf{D2C-Routing (full)} & 0.9846 & 0.8110 & 0.8995 & 0.6841 \\
Remove lexical-connective realization & 0.9773 & 0.7444 & 0.8860 & 0.5989 \\
Remove rhythm/POS & 0.9835 & 0.7922 & 0.8943 & 0.6287 \\
Remove surface-regularity & 0.9828 & 0.7878 & 0.8911 & 0.6177 \\
Lexical-connective realization only & 0.9808 & 0.7534 & 0.8809 & 0.5578 \\
Rhythm/POS only & 0.9825 & 0.7567 & 0.8830 & 0.5638 \\
Surface-regularity only & 0.9829 & 0.7854 & 0.8942 & 0.6125 \\
\bottomrule
\end{tabular}%
}
\end{table}

\begin{table}[!tbp]
\centering
\scriptsize
\setlength{\tabcolsep}{2pt}
\renewcommand{\arraystretch}{1.00}
\caption{Composition alternatives.}
\label{tab:composition-alternatives}
\resizebox{\columnwidth}{!}{%
\begin{tabular}{lrrrrr}
\toprule
Composition & AUROC & Avg TPR@1 & Macro-F1 & AH TPR@1 & AA TPR@1 \\
\midrule
\textbf{Learned vector gate} & 0.9846 & 0.8110 & 0.8995 & 0.7435 & 0.6841 \\
Fixed average of member models & 0.9805 & 0.7754 & 0.8880 & 0.7010 & 0.6072 \\
Concat without gate & 0.9814 & 0.7942 & 0.8959 & 0.7141 & 0.6846 \\
Hard factorized composition & 0.9810 & 0.7868 & 0.8971 & 0.7092 & 0.6213 \\
\bottomrule
\end{tabular}%
}
\end{table}
\FloatBarrier

Removing both margin and ranking losses remains close to the full objective, with 0.8146 Avg TPR@1\%FPR versus 0.8206 in the corresponding seed-42 setting; class-wise hard-negative OVR also does not improve AA TPR@1\%FPR. We therefore use ranking losses as evaluation-aligned optimization.

\subsection{Mechanism Diagnostics}\label{mechanism-diagnostics}

The supervised dimension heads are strong on their intended binary targets. On the MixD2C test split, the content head reaches AUROC 0.9937, accuracy 0.9653, and F1 0.9655 for AH/AA versus HH/HA. The expression head reaches AUROC 0.9871, accuracy 0.9606, and F1 0.9720 for HA/AA versus HH/AH. These results confirm that the intermediate source judgments are learnable under direct supervision.

Gate values themselves provide little explanatory signal. Their means are close to 0.5 across labels, ranging from 0.4975 for AA to 0.5067 for HH, and differ only slightly between correct and incorrect predictions. The supervised dimension heads and class-specific error patterns provide the more direct diagnostic evidence.

Scalar and vector learned gates are statistically indistinguishable at the ensemble level: vector/scalar Macro-F1 is 0.9077/0.9154 and Avg TPR@1 is 0.8235/0.8290, with paired intervals crossing zero. Together with the fixed-average, ungated-concat, and hard-factorized rows in Table~\ref{tab:composition-alternatives}, this result indicates that learned composition is stable across gate parameterizations; individual gate values should not be interpreted causally.

\begin{table}[!tbp]
\centering
\scriptsize
\setlength{\tabcolsep}{4pt}
\renewcommand{\arraystretch}{1.00}
\caption{Three-seed branch diagnosis for the two AI-content classes.}
\label{tab:ah-branch-diagnosis}
\resizebox{\columnwidth}{!}{%
\begin{tabular}{lrrrr}
\toprule
Class & Content acc. & Expression acc. & Correct expr. prob. & Four-way acc. \\
\midrule
AH & 0.9477 & 0.6438 & 0.6374 & 0.6634 \\
AA & 0.9721 & 0.9988 & 0.9958 & 0.9723 \\
\bottomrule
\end{tabular}
}
\end{table}

The AH gap is concentrated in expression-origin recognition: the model usually recognizes AI-originated content but often misses humanized expression. Two targeted expression-head remedies failed the predeclared development criterion and were not test-evaluated, leaving AH as the main single-model limitation; the 0.7892 system-level AH result is reported separately.

Length-bucket analysis shows a practical failure mode. Short texts are substantially harder, with Macro-F1 0.8633 and Avg TPR@1\%FPR 0.6930 in the shortest quartile, compared with Macro-F1 0.9303 and Avg TPR@1\%FPR 0.8927 in the longest quartile. This suggests that mixed-origin detection benefits from enough discourse and entity evidence to support the content-expression decomposition.

\subsection{Implementation}\label{appendix-implementation}

Training uses AdamW, learning rate \(1\times10^{-5}\), weight decay 0.01, 6\% linear warmup, five epochs, dropout 0.1, maximum sequence length 512, mixed precision, max gradient norm 1.0, and class weights for AH imbalance. Base dual-encoder runs use batch size 4 with gradient accumulation 8; shared-base and entity-as-text runs use batch size 8 with gradient accumulation 4, giving effective batch size 32.

D2C-base Fusion selects interpolation weights on the development split using an AH/AA/F1-weighted low-FPR objective, then evaluates once on test. The selected weights are 0.3733 for dual encoder seed44 with AH-over-AA margin, 0.4391 for dual encoder seed42 with strengthened expression supervision, and 0.1875 for entity-as-text seed42, so online inference uses three nonzero-weight member forward passes. For the group-aware robustness check only, the calibrated fusion system redistributes probability mass within the AH/AA pair without directly moving HH or HA probabilities.

SpecDetect uses GPT-2 and GPT-2 XL proxy language models; Fast-DetectGPT analytic uses GPT-2 XL as both scoring and reference model; the Binoculars proxy uses GPT-2 XL as observer and GPT-2 as performer, rather than the official Falcon-7B/Falcon-7B-Instruct pair. RACE-local uses the released RACE GNN with isanlp\_rst\_v3 rstdt parsing, 20 epochs, batch size 16, and dev F1-score checkpoint selection. Experiments run with PyTorch 2.5.1 and Transformers 4.44.2 on an NVIDIA RTX 4070 SUPER.

D2C-base Fusion runs three nonzero-weight members online: entity-as-text seed42 and two dual-encoder D2C-Routing variants. Their summed nonzero-member parameter count is 625.34M, with approximate summed latency of 43.52 ms/sample on RTX 4070 SUPER using dummy 512-token inputs. These values are relative cost disclosures, not hardware-independent speed claims.

\section{AI Assistance Disclosure}\label{appendix-ai-assistance}

We used AI assistance for grammar-level polishing and readability edits of the manuscript text. The research idea, experimental design, implementation, analysis, and final claims were reviewed and controlled by the authors.

\end{document}